\documentclass[journal]{IEEEtran}

\usepackage{graphics} 
\usepackage{epsfig} 
\usepackage{times} 
\usepackage{amsmath} 
\usepackage{amssymb}  
\usepackage{hyperref}
\usepackage{tabularx}
\usepackage{multirow}
\usepackage{makecell}
\usepackage{booktabs}
\usepackage{algorithmic}
\usepackage{graphicx}
\usepackage{xcolor}
\usepackage{cite}
\usepackage{hyperref}
\def\BibTeX{{\rm B\kern-.05em{\sc i\kern-.025em b}\kern-.08em
    T\kern-.1667em\lower.7ex\hbox{E}\kern-.125emX}}
\usepackage{balance}
\title{\LARGE \bf
UMCP: A Unified Multi-Task Collaborative Perception Network for Luggage Trolley Pose Estimation
}

\author{Zhirui Sun, Zhihao Jiang, Yao Wang, Jianwei Peng, and Jiankun Wang, \emph{Senior Member, IEEE} 
\thanks{This work is supported by National Natural Science Foundation of China under Grant 62473191, Shenzhen Key Laboratory of Robotics Perception and Intelligence (ZDSYS20200810171800001), Shenzhen Science and Technology Program under Grant 20231115141459001, RCBS20221008093305007, Guangdong Basic and Applied Basic Research Foundation under Grant 2025A1515012998, Young Elite Scientists Sponsorship Program by CAST under Grant 2023QNRC001, and the High level of special funds (G03034K003) from Southern University of Science and Technology, Shenzhen, China. \emph{(Corresponding author: Jiankun Wang).}}
\thanks{Zhirui Sun, Yao Wang, Jianwei Peng and Jiankun Wang are with Shenzhen Key Laboratory of Robotics Perception and Intelligence, Department of Electronic and Electrical Engineering, Southern University of Science and Technology, Shenzhen, China (e-mail: \url{wangjk@sustech.edu.cn}).}%
\thanks{Zhirui Sun, Zhihao Jiang, and Jiankun Wang are with Jiaxing Research Institute, Southern University of Science and Technology, Jiaxing, China.}%
}

\begin{document}

\maketitle
\thispagestyle{empty}
\pagestyle{empty}

\begin{abstract}
In robotic autonomous luggage trolley collection, robots must continuously localize scattered luggage trolleys in cluttered and dynamic environments. This requires the vision system to achieve both high accuracy and real-time performance. However, existing visual perception approaches for luggage trolleys often rely on cascaded multi-model inference, leading to increased inference latency and high deployment costs. To address these limitations, this article presents a unified multi-task collaborative perception network (UMCP) that simultaneously performs luggage trolley detection, keypoint detection and orientation estimation.
Based on the YOLOv12 architecture, keypoint features are fused with orientation features and then fed into an orientation feature enhancement module (OFEM), thereby improving orientation estimation accuracy. In addition, circular probability distribution modeling with a Kullback-Leibler (KL) divergence loss is adopted to enhance orientation estimation accuracy further.  Experimental results demonstrate that the proposed method achieves competitive overall accuracy while substantially reducing model complexity and computational cost compared with existing methods.
A website about this work is available at https://sites.google.com/view/robot-umcp.

\begin{IEEEkeywords}
Luggage trolley visual perception, Unified network, KL divergence loss.
\end{IEEEkeywords}


\end{abstract}


\section{Introduction}
\begin{figure}[t]
    \centering
    \includegraphics[width=\linewidth]{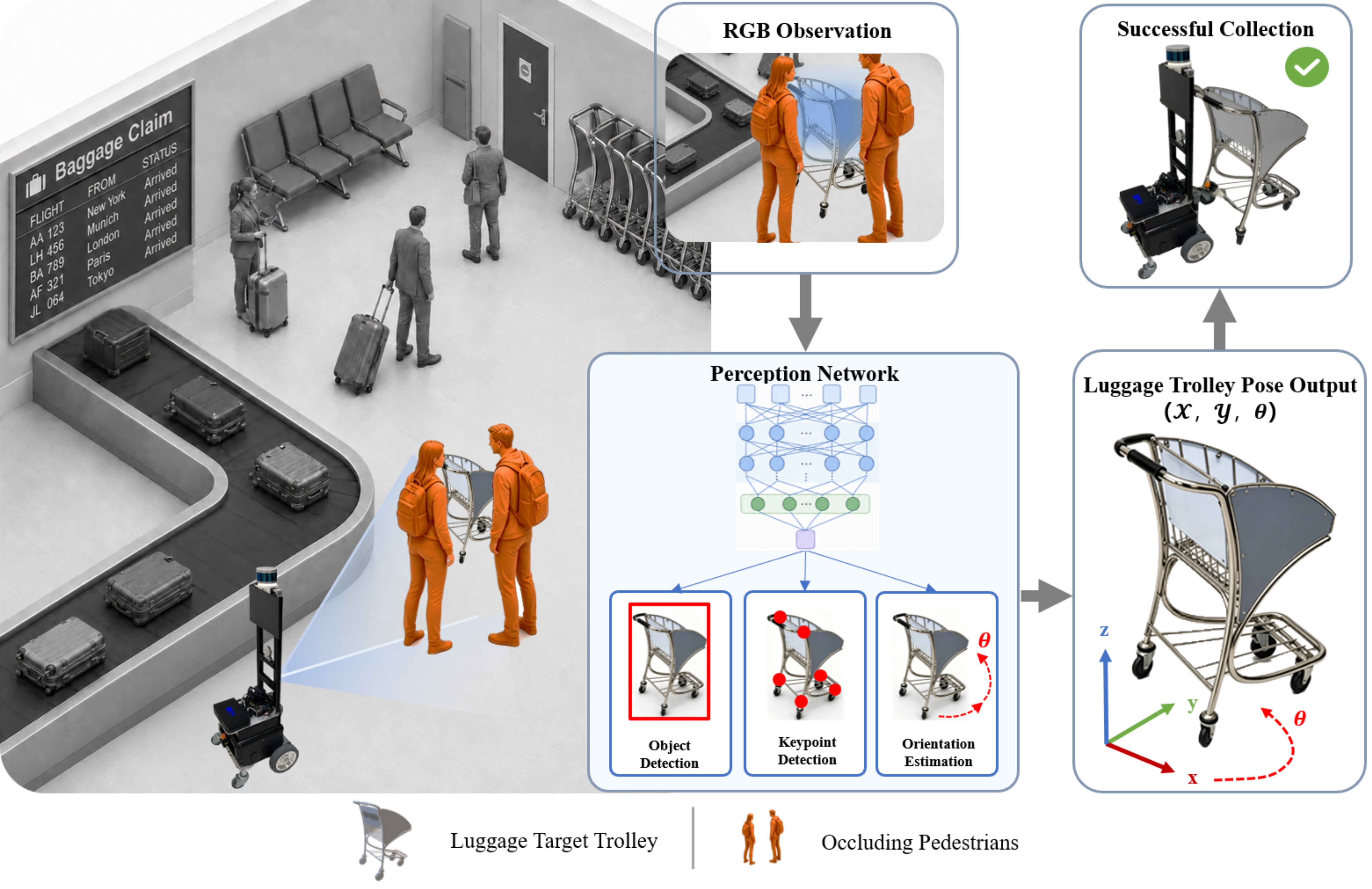}
    \caption{Schematic diagram of the proposed  robotic luggage trolley collection system.}
    \label{fig:show_image}
\end{figure}

\IEEEPARstart{R}{obotic} systems for luggage trolley collection require the tight integration of visual perception, motion planning and control. Visual perception provides the pose of the luggage trolley for downstream planning and control, motion planning generates feasible, collision-free trajectories and control ensures accurate execution.  As illustrated in Fig. ~\ref{fig:show_image}, two people stopping to talk may partially occlude an idle luggage trolley, making its detection and localization more challenging. Therefore, robust detection and localization under partial occlusion are crucial for the system, as they enable the robot to accurately estimate the trolley pose, which forms the foundation for subsequent motion planning and precise end-effector manipulation. Although existing studies~\cite{xiao2022robotic,xie2024autonomous,sun2023evaluation} have advanced the system from different perspectives, reliable deployment in real-world scenarios remains difficult, particularly in large-scale and dynamic environments such as airports~\cite{sun2024uwb}. In these environments, unpredictable human activities, cluttered surroundings and frequent occlusions introduce substantial uncertainty, making accurate perception significantly more challenging. Since errors in perception can directly cause navigation drift and grasping failures, visual perception becomes a critical bottleneck in overall system.

Some methods estimate the pose of luggage trolleys by solving EPnP~\cite{lepetit2009epnp} using detected keypoints. Although effective under ideal conditions, these approaches depend strongly on the accuracy and completeness of the detected keypoints. Under partial occlusion, missing keypoints may cause these methods to fail. This limitation is particularly critical in crowded environments, where occlusion occurs frequently and reliable keypoint detection cannot be guaranteed. To improve robustness under occlusion, some studies estimate pose through point cloud registration~\cite{pan2020searching}. Although these methods reduce reliance on complete keypoint visibility, they often remain inadequate in terms of real-time performance and accuracy~\cite{zheng2023refusion}.
In addition, point-cloud-based methods typically require expensive sensors or dense computations, which limit their applicability on resource-constrained robotic platforms. The approach by Sun \textit{et al.}~\cite{sun2025hp} employs a hierarchical progressive perception system for luggage trolley detection and localization (referred to as HPPS). It constructs a geometric model and estimates the pose using a few visible keypoints. However, the pipeline relies on multiple networks for object detection, keypoint detection and orientation estimation, resulting in higher computational cost and a larger parameter count. Furthermore, deploying independent modules sequentially compounds inference latency and propagates cumulative errors across stages, which is highly detrimental to the real-time responsiveness required in dynamic, human-populated environments. Therefore, there is a pressing need for a lightweight, unified perception framework capable of handling these multiple tasks simultaneously without compromising accuracy~\cite{song2025fislam}.

To address the above issues, we propose UMCP.
This network jointly performs object detection, keypoint detection and orientation estimation, enabling collaborative feature learning while reducing the number of model parameters.
Built upon the YOLOv12~\cite{tian2025yolov12} backbone, we introduce an additional orientation estimation branch. The backbone provides strong feature extraction capabilities. To improve orientation estimation, keypoint features are fused with orientation features and an OFEM is used to enhance feature representations. We further incorporate a KL divergence loss for orientation estimation.
In contrast to Gaussian regression methods such as MEBOW~\cite{wu2020mebow}, we directly enforce consistency between the predicted and ground-truth distributions via KL divergence, thereby avoiding instability caused by explicit distribution-parameter regression and improving convergence during training. Collectively, these components integrate object detection, keypoint detection and orientation estimation within a unified network. Finally, the 3D coordinates of the luggage trolley are computed using camera projection geometry, providing a basis for precise collection.

The contributions of this article are summarized as follows:
\begin{itemize}
    \item This article proposes UMCP, a unified multi-task perception network that jointly performs luggage trolley detection, keypoint detection and orientation estimation, thereby reducing redundant parameters while improving inference efficiency.
    \item This article introduces a lightweight OFEM for improved orientation estimation, together with a KL divergence loss that aligns the predicted and ground-truth orientation distributions, resulting in more stable and accurate predictions.
    \item Extensive experiments demonstrate that the proposed method achieves competitive overall performance across the three tasks while maintaining a compact model size and low computational complexity, making it suitable for practical deployment.
\end{itemize}

\section{Related Work}
This section reviews prior work related to object detection, keypoint detection and orientation estimation. Despite recent advances, achieving both robustness and efficiency in luggage trolley pose estimation under cluttered, occluded conditions remains challenging.
\subsection{Object Detection}
Object detection is a fundamental task in visual perception. Early methods primarily relied on sliding-window mechanisms and handcrafted feature design~\cite{viola2001rapid,dalal2005hog}, which not only incurred high computational costs but also lacked robustness to complex backgrounds and scale variations. With the advancement of deep learning, detection frameworks have gradually evolved into end-to-end convolutional neural network models~\cite{krizhevsky2012imagenet}, significantly improving detection accuracy. The R-CNN series~\cite{girshick2016rcnn, ren2015fasterrcnn} substantially advanced detection performance by introducing region proposal mechanisms and deep feature representations. However, these two-stage methods usually require repeated feature extraction and classification over a large number of proposals, resulting in high computational cost and making real-time applications challenging. Although Fast R-CNN and Faster R-CNN improve efficiency through feature sharing and the introduction of the Region Proposal Network (RPN), the overall inference process remains relatively complex, posing challenges for deployment on resource-constrained robotic platforms. Mask R-CNN further extends these methods to instance segmentation, enhancing model expressiveness while also introducing greater computational overhead.

To satisfy real-time requirements, one-stage detectors have gradually become a major research focus. The YOLO series~\cite{redmon2016yolo} formulates object detection as a regression problem, predicting object classes and bounding boxes within a single network, thereby substantially improving inference speed. However, early versions showed limitations in small-object detection and localization accuracy. SSD~\cite{liu2016ssd} mitigates the issue of small object detection by employing multi-scale feature maps, but it remains susceptible to class imbalance in complex scenarios. To further improve multi-scale feature representation, feature fusion architectures such as FPN~\cite{lin2017fpn}, PANet~\cite{liu2018path}, and BiFPN~\cite{tan2020efficientdet} have been widely integrated into detection frameworks, significantly enhancing detection performance across different scales. However, these methods often rely on complex feature fusion designs, which increase network complexity while improving accuracy~\cite{zheng2023refusion}. 

\subsection{Keypoint Detection}
Keypoint detection enables the precise localization of specific object parts in images. Early methods, such as DeepPose~\cite{toshev2014deeppose}, directly formulated keypoint detection as a regression problem. Although structurally simple, these approaches exhibit limited robustness to complex backgrounds and occlusions. To enhance spatial modeling capability, Stacked Hourglass Networks~\cite{newell2016stacked} employ repeated multi-scale feature encoding and decoding, thereby effectively improving local structural reasoning. However, their stacked architecture incurs a high computational cost, which is unfavorable for real-time applications. HRNet~\cite{wang2020hrnet} preserves high-resolution feature representations throughout the network, enabling the simultaneous modeling of fine-grained structural information and global context and achieves state-of-the-art performance on keypoint detection benchmarks such as COCO.
Nevertheless, its multi-branch, high-resolution design substantially increases the number of model parameters and computational complexity, making deployment on embedded or mobile platforms challenging. To improve robustness under occlusion, CoKe~\cite{bai2023coke} introduces a contrastive learning mechanism to enhance the discriminability of keypoint features, but its complex training procedure requires high-quality data. 
Furthermore, Transformer-based architectures have also been applied to keypoint detection~\cite{shi2022end}, enabling end-to-end prediction through global attention. 
However, they likewise suffer from high computational overhead and limited real-time performance.

YOLO-KP~\cite{maji2022yolopose} integrates keypoint detection into a single-stage detection framework, thereby offering significant advantages in inference efficiency. However, owing to limited feature resolution, its accuracy under complex poses and severe occlusions remains improvable. OpenPose~\cite{Openpose} adopts a bottom-up multi-person keypoint detection strategy suitable for human pose analysis, but its complex network architecture impedes direct transfer to specific industrial applications. 

\subsection{Orientation Estimation}
Orientation estimation is a fundamental component of autonomous robotic luggage trolley collection. With the development of deep learning, orientation estimation has increasingly been formulated as either a classification or a regression problem, achieving higher accuracy through end-to-end training. For instance,~\cite{raza2018appearance} employs a multilayer convolutional network to predict a limited set of discrete orientation bins. Although such methods are structurally simple and stable during training, discretization inevitably introduces quantization errors, making high-precision continuous angle estimation difficult. Direct regression of continuous angles avoids discretization errors, yet the inherent periodicity of angles can cause boundary discontinuities, leading to unstable training. MEBOW~\cite{wu2020mebow} addresses the periodicity issue by modeling orientation as a fixed circular Gaussian distribution. However, under conditions of limited occluded samples, regression-based strategies tend to produce low-confidence outputs that are difficult to distinguish. Yu \textit{et al.} ~\cite{yu2019continuous} and Part-HOE~\cite{zhao2024parthoe} exploit spatial geometric relationships among keypoints or leverage visible target information to assist orientation estimation, demonstrating greater robustness under occlusion. Nevertheless, these approaches generally involve complex networks with many parameters, making them difficult to integrate into lightweight, real-time detection frameworks~\cite{hu2025amalign}.

\section{Methodology}
This section introduces the components of the proposed UMCP and their implementation details. The following subsections describe the overall network framework, the multi-task object and keypoint detection head, the 3D coordinate computation process, the orientation estimation module and the KL divergence loss function. Vectors are denoted by boldface letters.
\subsection{Overall Network Architecture}\label{AA}
\begin{figure*}[t]
    \centering
    \includegraphics[width=\linewidth]{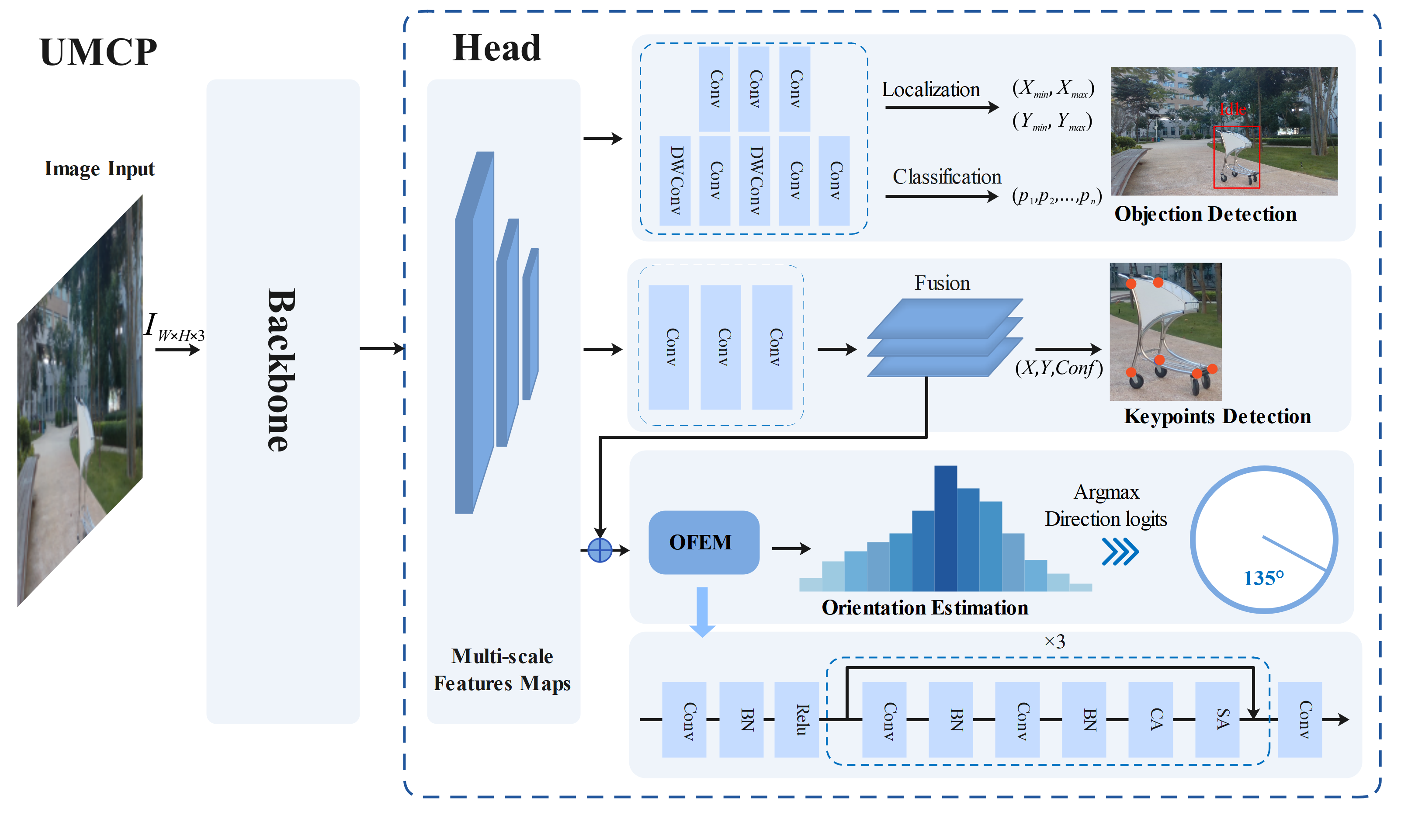}
    \caption{Overall architecture of the proposed UMCP. Given an input RGB image, a shared backbone extracts multi-scale feature representations, which are then fed into a detection head, a keypoint head and an orientation estimation head. The orientation estimation head fuses keypoint features, is further enhanced by an OFEM and is optimized using a KL divergence loss function to achieve accurate orientation estimation.}
    \label{fig:architecture}
\end{figure*}

The proposed UMCP is designed as a unified multi-task framework that jointly performs object detection, keypoint detection and orientation estimation. Given an input RGB image, the backbone network first extracts multi-scale feature representations, producing semantic feature maps. These features are subsequently fed into the corresponding heads. Based on the detected keypoints, the 3D coordinates of the luggage trolley are finally computed using camera projection geometry. To facilitate accurate orientation estimation, an additional orientation feature branch is introduced during feature extraction. The orientation features are fused with the keypoint features, effectively integrating global semantic information with geometric structure derived from keypoints. Furthermore, an OFEM is incorporated to strengthen orientation-related features by adaptively focusing on regions that are highly informative for orientation estimation. The overall network architecture is illustrated in Fig.~\ref{fig:architecture}.

\subsection{Multi-task Object and Keypoint Detection}
Given an input RGB image, the proposed UMCP employs a unified network to detect luggage trolleys and keypoints jointly. The object detection branch predicts the bounding box of the luggage trolley, while the keypoint detection branch outputs the 2D image coordinates of the luggage trolley keypoints.
Based on the detected keypoints, a prior geometric model of the luggage trolley is further utilized to estimate the 3D coordinates. Specifically, the center of the luggage trolley is assumed to lie on the ground plane and the relative spatial configuration of each keypoint with respect to the center is known a priori.
The geometric prior of the luggage trolley is defined as follows:
\begin{equation}
\mathcal{P} = \left\{ \mathbf{p}_i = (X_i, Y_i, H_i) \in \mathbb{R}^3 \mid i = 0,1,\ldots,5 \right\},
\end{equation}
where $\mathbf{p}_i$ denotes the 3D offset vector of the $i$-th keypoint relative to the center of the luggage trolley and consists of the planar components $X_i$, $Y_i$ and the vertical height $H_i$. Assuming a known ground plane with normal vector $\mathbf{N} \in \mathbb{R}^3$ and camera height $h$, a point $\mathbf{x}\in\mathbb{R}^3$ in the camera coordinate frame lying on the ground plane satisfies:
\begin{equation}
\mathbf{-N}^{\top} \mathbf{x} +h= 0.
\end{equation}

Let $\hat{\boldsymbol{\alpha}}_i$ denote the homogeneous image coordinates of the $i$-th keypoint predicted by the network and let $\mathbf{K}$ denote the camera intrinsic matrix. The back-projected ray corresponding to the $i$-th keypoint in the camera coordinate frame is given by $\mathbf{K}^{-1}\hat{\boldsymbol{\alpha}}_i$.
The 3D coordinates of the keypoint are obtained by intersecting this ray with the plane parallel to the ground plane at height $H_i$:
\begin{equation}
\mathbf{x}_i = \lambda_i \mathbf{K}^{-1}\hat{\boldsymbol{\alpha}}_i,
\qquad
\lambda_i = \frac{h - H_i}{\mathbf{N}^{\top}\mathbf{K}^{-1}\hat{\boldsymbol{\alpha}}_i},
\end{equation}
where $\lambda_i$ denotes the scale factor along the back-projected ray, such that the reconstructed keypoint lies on the plane parallel to the ground plane with height offset $H_i$. To ensure positive depth in the camera coordinate frame, the sign of $\lambda_i$ is constrained during implementation.

Using the reconstructed 3D coordinates of all visible keypoints, the luggage trolley center is estimated as the weighted mean of the individual keypoint estimates.
\begin{equation}
\mathbf{C} = \frac{1}{\sum_{i \in \mathcal{V}} w_i}
\sum_{i \in \mathcal{V}} w_i \, (\mathbf{x}_i-\mathbf{p}_i),
\end{equation}
where $\mathbf{C}$ denotes the estimated 3D center of the luggage trolley, $\mathcal{V}$ is the set of visible keypoints and $w_i$ represents the confidence score of the $i$-th keypoint predicted by the network. This weighted formulation enables the model to exploit all available keypoints while assigning greater influence to more reliable predictions, thereby improving  accuracy.

\subsection{Orientation Feature Enhancement Module}
We introduce an OFEM to enhance orientation estimation accuracy by strengthening orientation-sensitive feature representations. The OFEM employs a residual learning framework integrated with a Convolutional Block Attention Module, which combines channel and spatial attention mechanisms to emphasize discriminative features while selectively preserving the original feature resolution.

Given an input feature map $\mathbf{X} \in \mathbb{R}^{C \times H \times W}$, the OFEM first applies a channel attention mechanism to recalibrate channel-wise feature responses adaptively. A global average pooling operation is performed over the spatial dimensions to obtain a channel-wise descriptor:
\begin{equation}
z_c=\frac{1}{HW}\sum_{i=1}^{H}\sum_{j=1}^{W} X_{c,i,j},
\end{equation}
where $z_c$ denotes the aggregated response of the $c$-th channel. The descriptor vector $\mathbf{z}=[z_1,\dots,z_c]$ is subsequently passed through two fully connected layers, implemented as convolutions, with a ReLU activation function in between to model inter-channel dependencies:
\begin{equation}
\mathbf{s}=\sigma \left(W_2\,\mathrm{ReLU}(W_1 \mathbf{z})\right),
\end{equation}
where $\sigma(\cdot)$ denotes the sigmoid activation function and $W_1$ and $W_2$ are learnable parameters. The input feature map is then recalibrated using the resulting channel attention weights as follows:
\begin{equation}
\mathbf{X}^{\mathrm{ca}}_c = s_c \cdot \mathbf{X}_c,
\end{equation}
where $\mathbf{X}^{\mathrm{ca}}$ denotes the feature map after the application of channel attention.

Following channel attention, a spatial attention mechanism is employed to further enhance feature selectivity across spatial dimensions. Given $\mathbf{X}^{\mathrm{ca}}$, channel-wise average pooling and max pooling are applied to generate two spatial descriptors:
\begin{equation}
F_{\mathrm{avg}}=\mathrm{AvgPool}(\mathbf{X}^{\mathrm{ca}}), \quad
F_{\mathrm{max}}=\mathrm{MaxPool}(\mathbf{X}^{\mathrm{ca}}),
\end{equation}
where both $F_{\mathrm{avg}}$ and $F_{\mathrm{max}}$ are spatial maps of size $H \times W$. These two maps are concatenated along the channel dimension and then passed through a convolutional layer followed by a sigmoid activation function to generate a spatial attention map. The spatially refined feature map is expressed as:
\begin{equation}
\mathbf{X}^{\mathrm{sa}} = \sigma\left(\mathrm{Conv}\big([F_{\mathrm{avg}};F_{\mathrm{max}}]\big)\right)
\odot \mathbf{X}^{\mathrm{ca}},
\end{equation}
where $\odot$ denotes element-wise multiplication and $\mathbf{X}^{\mathrm{sa}}$ denotes the spatially attended features.
To promote stable gradient propagation and deeper feature learning, the OFEM employs a residual connection:
\begin{equation}
\mathbf{Y} = \mathbf{X} + \mathbf{X}^{\mathrm{sa}}.
\end{equation}

Finally, a convolutional layer is employed to project the output features onto the desired number of channels. The overall formulation of the OFEM can be expressed as:
\begin{equation}
\mathrm{OFEM}(\mathbf{X})=
\mathrm{Conv}\big(\mathbf{X}+\mathrm{SA}(\mathrm{CA}(\mathbf{X}))\big),
\end{equation}
where $\mathrm{CA}(\cdot)$ and $\mathrm{SA}(\cdot)$ denote the channel attention and spatial attention operations, respectively.

\subsection{KL Loss Function}
The detection loss and keypoint loss follow the standard formulation, whereas we focus on the orientation loss.
We adopt a Gaussian-based formulation that models orientation estimation as a probability distribution defined on a circular angular domain.
Specifically, the $360^\circ$ orientation range is uniformly divided into $C$ discrete bins, each spanning an angle of $360^\circ/C$ degrees.
In our implementation, we set $C=360$, such that each bin corresponds to one degree. To account for the periodic nature of orientation, we define the circular distance between the $i$-th orientation bin and the ground-truth bin $l_{\mathrm{gt}}$ is defined as:
\begin{equation}
\Delta_{i,l_{\mathrm{gt}}}
=\min\!\left(
\left| i-l_{\mathrm{gt}} \right|,\;
C-\left| i-l_{\mathrm{gt}} \right|
\right),
\end{equation}
where $\Delta_{i,l_{\mathrm{gt}}}$ denotes the minimum angular distance on the circular domain.
Let $i \in \{0, \dots, C-1\}$ denote the index of an orientation bin.

Based on this distance, we define a circular Gaussian weighting function centered at the ground-truth orientation:
\begin{equation}
g_i
=\exp\!\left(
-\frac{\Delta_{i,l_{\mathrm{gt}}}^{2}}{2\sigma^{2}}
\right),
\end{equation}
where $\sigma$ is a hyperparameter representing the standard deviation and controlling the smoothness of the distribution. The Gaussian weighting assigns higher values to bins closer to the ground-truth orientation and gradually decreases for bins farther away, thereby providing smooth, continuous supervision around the true angle. To obtain a valid probability distribution, we normalize the Gaussian weights to construct a circular Gaussian target distribution:
\begin{equation}
\varphi_i
=\frac{g_i}{\sum_{j=0}^{C-1} g_j}.
\end{equation}

We quantify the discrepancy between the predicted distribution and the target distribution using the KL divergence loss:
\begin{equation}
L_{\mathrm{yaw}}
=\frac{1}{B}
\sum_{b=1}^{B}
\sum_{i=0}^{C-1}
\varphi_{b,i}
\log\frac{\varphi_{b,i}}{p_{b,i}},
\end{equation}
where $B$ denotes the batch size and $p_{b,i}$ represents the predicted probability that the $b$-th sample belongs to the $i$-th orientation bin.
 
Since the target distribution $\varphi_{b,i}$ is independent of the network parameters, minimizing the KL divergence is equivalent, up to an additive constant, to minimizing a cross-entropy loss with soft labels.
Thus, the final orientation loss is given by:
\begin{equation}
L_{\mathrm{yaw}}
=-\frac{\lambda}{B}
\sum_{b=1}^{B}
\sum_{i=0}^{C-1}
\varphi_{b,i}\log p_{b,i},
\end{equation}
where $\lambda$ is a hyperparameter controlling the relative contribution of the yaw estimation loss.
The overall training objective is then formulated as the weighted sum of the yaw estimation loss, the keypoint detection loss, and the object detection loss:
\begin{equation}
L_{\mathrm{total}} = \alpha L_{\mathrm{yaw}} + \beta L_{\mathrm{kpt}} + \gamma L_{\mathrm{det}},
\label{eq:total_loss}
\end{equation}
where $\alpha$, $\beta$, and $\gamma$ are hyperparameters. Specifically, $L_{\mathrm{kpt}}$ denotes the keypoint detection loss and $L_{\mathrm{det}}$ denotes the object detection loss. In our experiments, we set $\alpha = 0.01$, $\beta = 1.0$ and $\gamma = 1.0$.
 
\section{Experiments}
\begin{figure}[t]
    \centering
    \includegraphics[width=\linewidth]{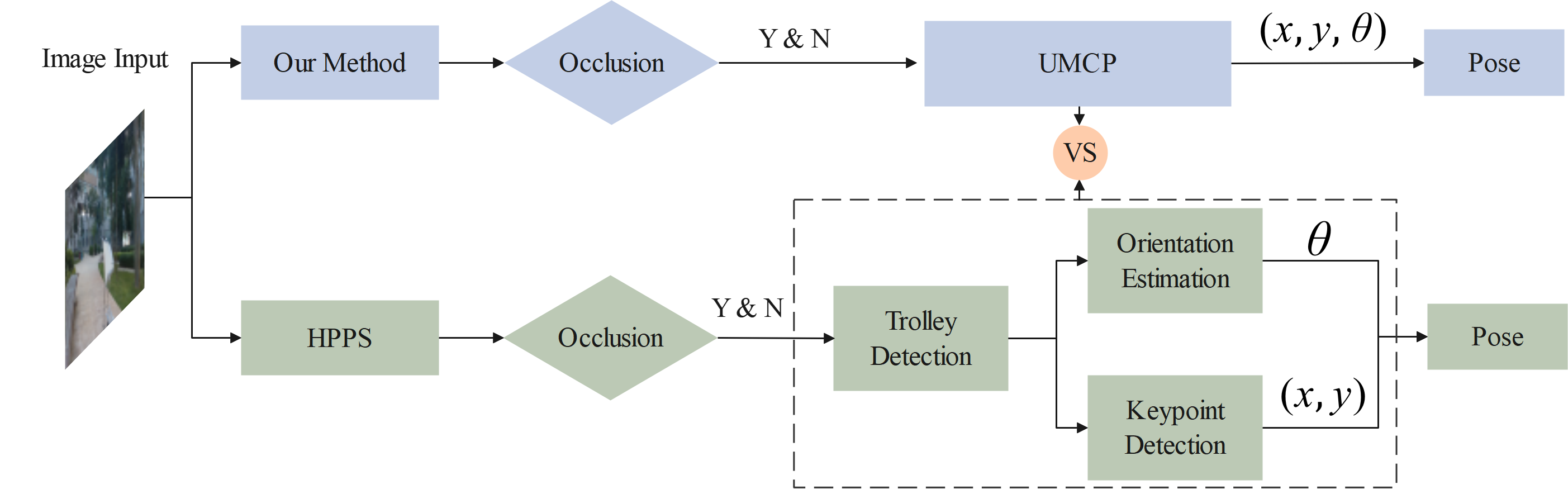}
     \caption{Comparison of the workflows of our method and HPPS for luggage trolley pose estimation from image inputs.}
     \label{fig:overnet}
\end{figure}

This section presents the implementation details, experimental results and ablation studies.
To comprehensively evaluate the proposed method, we compare it with HPPS~\cite{sun2025hp}, which achieves strong performance in luggage trolley perception tasks, particularly under occluded conditions.
Fig.~\ref{fig:overnet} illustrates the experimental workflow and contrasts our method with HPPS for luggage trolley pose estimation from image inputs. Our method jointly performs object detection, keypoint detection and orientation estimation within a unified network, enabling collaborative feature learning for pose estimation. In contrast, HPPS decouples object detection, keypoint detection and orientation estimation. For a clear comparison, we focus on three critical tasks: object detection, keypoint detection and orientation estimation. 
\subsection{Implementation Details}
All experiments are conducted on the luggage trolley dataset proposed by HPPS~\cite{sun2025hp}, which is designed for luggage trolley detection, keypoint detection and orientation estimation. The dataset encompasses diverse luggage trolley types and complex scenes. Each instance is annotated with a 2D bounding box, six keypoints with visibility labels and an orientation angle. We randomly split the dataset into $80\%$ for training and $20\%$ for testing.

Our method is trained offline using PyTorch on a platform equipped with an AMD EPYC MILAN 7413 CPU and an NVIDIA RTX A6000 GPU.
We employ SGD as the optimizer and adopt a two-stage training strategy. In the first stage, object detection and keypoint detection are optimized. In the second stage, orientation estimation is further refined. An early stopping strategy is applied during training.

For the luggage trolley detection task, standard evaluation metrics, including Precision (P), Recall (R), mAP50 and mAP50–95, are employed, with higher values indicating better performance. For keypoint detection, PCK@0.01 is adopted as the evaluation metric. It measures the proportion of detected keypoints that lie within 1\% of the image diagonal from their ground-truth locations, with higher values indicating better accuracy. For orientation estimation, the Average  Angular Error (AAE) is used to quantify the average deviation between predicted and ground truth orientations, with lower values indicating better performance. In addition, to assess practical deployability, model complexity and efficiency are evaluated in terms of the number of parameters, Floating Point Operations (FLOPs), and latency.

\subsection{Experimental Results}
\subsubsection{Visualization Results}
\begin{figure*}[htbp]
    \centering
    \includegraphics[width=0.98\linewidth]{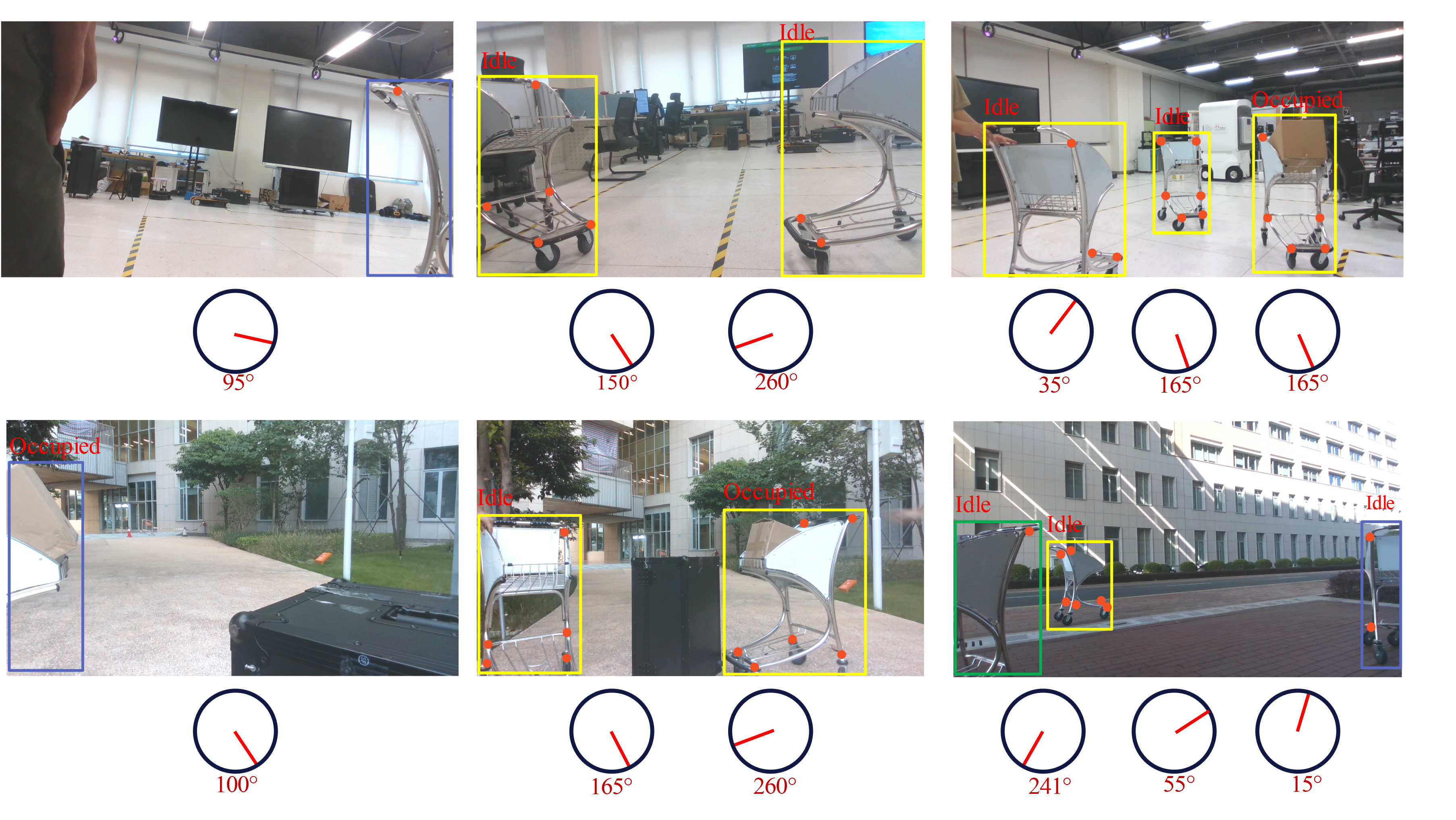}
   \caption{Results for indoor scenes are shown in the top row, while results for outdoor scenes are shown in the bottom row. 
    Yellow, green and blue indicate visibility levels of $>80\%$, $40\%-80\%$ and $<40\%$, respectively.}
     \label{fig:example}
\end{figure*}

\begin{figure}[t]
    \centering
    \includegraphics[width=\linewidth]{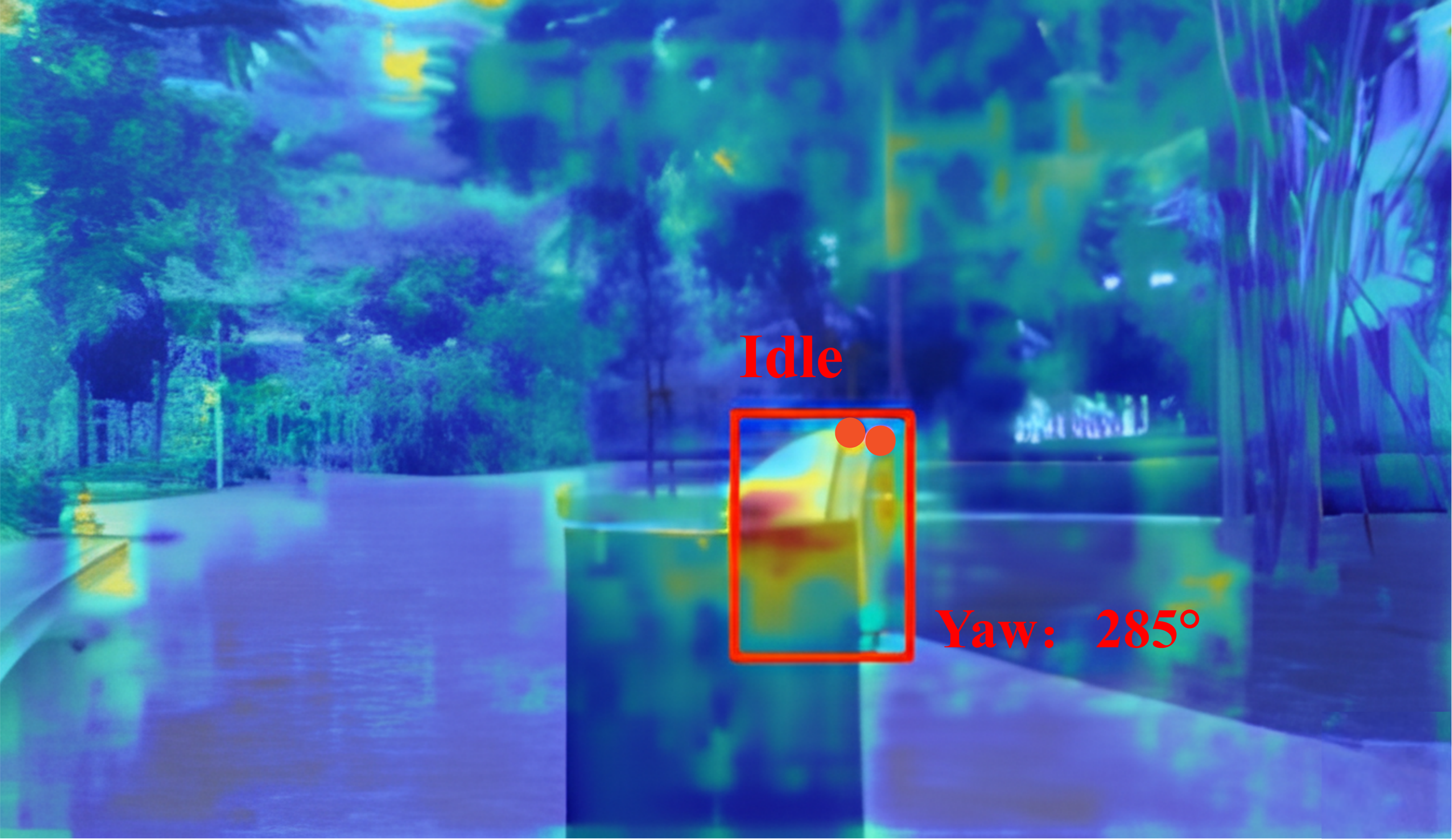}
     \caption{Visualization of multi-task prediction results in the form of heatmaps, including object detection, keypoint detection and orientation estimation}
     \label{fig:heatmap}
\end{figure}

Fig.~\ref{fig:example} visualizes the estimated luggage trolley positions, keypoints and orientations. The first row shows indoor examples, while the second row presents outdoor cases. To clarify the orientation estimation, the predicted yaw angles are plotted beneath each image, from left to right, corresponding to the orientations of the luggage trolleys in the image.
To further interpret the model behavior, heatmap-based visual explanations are provided. As shown in Fig.~\ref{fig:heatmap}, high-response regions are concentrated within the target bounding boxes, indicating that the model effectively focuses on features related to luggage trolleys while suppressing background interference.

\subsubsection{Object Detection Results}
The comparison of object detection performance is presented in Tab.~\ref{tab:det}. Based on the dataset, performance is evaluated from three perspectives. Visibility represents the extent to which a luggage trolley is observable in the image and is categorized into three levels: $> 80\%$, $40\%-80\%$ and $< 40\%$. Count denotes the number of luggage trolleys present in an image. State represents the loading condition of a luggage trolley and is classified into two categories: idle and occupied. Our method consistently achieves high detection accuracy across different scenarios, demonstrating strong robustness under varying levels of occlusion, target density and operational state. Under different visibility conditions, UMCP maintains stable precision and recall, especially under medium and low visibility ($40\%-80\%$ and $< 40\%$). Compared with HPPS, UMCP achieves a notably higher mAP50-95 in partially occluded scenarios, indicating improved localization accuracy when luggage trolleys are partially obstructed. In scenarios with varying target counts, UMCP consistently performs competitively.

While both methods perform well for a single target, UMCP shows greater advantages in multi-target settings, particularly when two or three luggage trolleys are present. These results suggest that the proposed unified framework generalizes effectively to crowded scenes without significant degradation in detection accuracy. Across different luggage trolley states, UMCP outperforms HPPS in the idle scenario and achieves comparable precision and recall in the occupied case.
In summary, the proposed method achieves competitive detection performance across most evaluated scenarios while substantially reducing model complexity.
\begin{table}[htbp]
\centering
\caption{Detection performance comparison under different scenarios}
\label{tab:det}
\resizebox{1\linewidth}{!}{%
\renewcommand{\arraystretch}{2.2}
\setlength{\tabcolsep}{3pt}
\begin{tabular}{c c ccccc ccccc}
    \toprule[1pt]
    \multirow{2}{*}{Metric} & \multirow{2}{*}{Scenario} &
    \multicolumn{4}{c}{HPPS~\cite{sun2025hp}} &
    \multicolumn{4}{c}{UMCP} \\
    \cline{3-10}
     &  & P & R & mAP50 & mAP50-95 & P & R & mAP50 & mAP50-95 \\
  \hline
        \multirow{3}{*}{Visibility}
    & $> 80\%$
    & 0.999 & 1.000 & 0.995 & 0.993
    & 0.999 & 1.000 & 0.995 & \textbf{0.995} \\
    
    & $40\%-80\%$
    & 0.998 & 1.000 & 0.995 & 0.975
    & \textbf{0.999} & 1.000 & 0.995 & \textbf{0.989} \\
    
    & $< 40\%$
    & \textbf{1.000} & 0.924 & \textbf{0.989} & 0.908
    & 0.946 & \textbf{0.960} & 0.983 & \textbf{0.927} \\
    \hline
    \multirow{3}{*}{Count}
    & One
    & \textbf{1.000} & 0.971 & \textbf{0.994} & 0.961
    & 0.988 & \textbf{0.979} & 0.993 & \textbf{0.975} \\
    
    & Two
    & \textbf{0.970} & 0.967 & 0.979 & 0.951
    & 0.964 & \textbf{0.988} & \textbf{0.993} & \textbf{0.975} \\
    
    & Three
    & 0.997 & 0.996 & 0.995 & 0.988
    & \textbf{0.998} & \textbf{0.999} & 0.995 & \textbf{0.993} \\
    \hline
    \multirow{2}{*}{State}
    & Idle
    & 0.991 & 0.988 & 0.994 & 0.983
    & \textbf{0.993} & \textbf{0.994} & 0.994 & \textbf{0.989} \\
    
    & Occupied
    & \textbf{0.995} & 0.990 & 0.994 & \textbf{0.982}
    & 0.994 & \textbf{0.994} & 0.994 & 0.930 \\
    \hline
     & Overall
    & 0.990 & 0.989 & 0.994 & 0.982
    & \textbf{0.994} & \textbf{0.994} & 0.994 & \textbf{0.989} \\
    \bottomrule [1pt]
\end{tabular}}
\end{table}

\subsubsection{Keypoint Detection Results}
Tab.~\ref{tab:kpt} presents the PCK@0.01 results for keypoint detection under different visibility conditions. Our method substantially outperforms HPPS under high-visibility settings, indicating superior localization accuracy when keypoints are clearly visible. Under medium- and low-visibility conditions, our approach maintains competitive performance without substantial degradation.Our method achieved the best performance on the entire dataset.
Overall, UMCP achieves the best aggregate PCK@0.01 and shows a clear advantage under high-visibility conditions, while remaining competitive under medium- and low-visibility settings.

\begin{table}[htbp]
\centering
\caption{Comparison of PCK@0.01 for keypoint detection}
\label{tab:kpt}
\resizebox{\linewidth}{!}{%
\renewcommand{\arraystretch}{1.5}
\setlength{\tabcolsep}{15pt}   
\begin{tabular}{c c c}
\toprule[1pt]
Visibility & HPPS~\cite{sun2025hp} & UMCP \\
\hline
$> 80\%$        & 0.793 & \textbf{0.931} \\
$40\%-80\%$    & \textbf{0.777} & 0.761 \\
$< 40\%$        & \textbf{0.652} & 0.579 \\
Overall        & 0.767 & \textbf{0.908} \\   
\bottomrule [1pt] 
\end{tabular}
}
\end{table}

\subsubsection{Orientation Estimation Results}
Tab.~\ref{tab:yaw} reports the AAE comparison of different methods for orientation estimation under varying visibility levels, target counts and luggage trolley states. Overall, the proposed method outperforms HPPS in most scenarios, particularly in complex environments. 
Across most evaluated conditions, our method achieves lower AAE, especially under occlusion and multi-target settings.

\begin{table}[htbp]
\centering
\caption{AAE comparison of orientation estimation under different scenarios}
\label{tab:yaw}
\resizebox{1\linewidth}{!}{%
\renewcommand{\arraystretch}{1.7}
\setlength{\tabcolsep}{13pt}
\begin{tabular}{c c c c}
\toprule[1pt]
Metric & Scenario & HPPS~\cite{sun2025hp} & UMCP \\
\hline
\multirow{3}{*}{Visibility}
& $>80\%$      & 6.75 & \textbf{5.91} \\
& $40\%-80\%$  & 7.44 & \textbf{6.78} \\
& $<40\%$      & 9.18 & \textbf{8.02} \\
\hline
\multirow{3}{*}{Count}
& One   & 7.62 & \textbf{7.10} \\
& Two   & 8.67 & \textbf{6.98} \\
& Three & 7.38 & \textbf{6.49} \\
\hline
\multirow{2}{*}{State}
& Occupied & 8.23 & \textbf{6.70} \\
& Idle    & \textbf{8.53} & 9.31 \\
\hline
 & Overall & 8.19 & \textbf{7.76} \\
\bottomrule [1pt]
\end{tabular}
}
\end{table}

\subsubsection{Model Complexity and Network Scale}
Tab.~\ref{tab:speed} compares different methods in terms of model size and computational cost. O, K and Y denote object detection, keypoint detection  and orientation estimation, respectively. For HPPS, the reported total complexity corresponds to the sum of its two separate networks. As shown in Tab. ~\ref{tab:speed}, UMCP jointly performs all three tasks within a unified network, requiring only 3.62 M parameters and 5.38 G FLOPs. In contrast, HPPS relies on a two-stage design, resulting in a total of 42.45 M parameters and 79.88 G FLOPs when all tasks are considered. This corresponds to a reduction of approximately 91\% in the number of parameters and 93\% in FLOPs compared with HPPS. Despite the substantially lower computational complexity, UMCP maintains competitive performance across all tasks.   
These results demonstrate that the proposed unified network achieves substantially lower model complexity while maintaining competitive performance.

\begin{table}[htbp]
\centering
\caption{Comparison of computational complexity}
\label{tab:speed}
\resizebox{1\linewidth}{!}{%
\renewcommand{\arraystretch}{1.5}  
\setlength{\tabcolsep}{10pt} 
\begin{tabular}{c c c c c}
\toprule[1pt]  
Method & Task  & Param (M) & FLOPs (G)&Latency(ms) \\
\hline
\multirow{3}{*}{HPPS\cite{sun2025hp}} & O & 2.82 & 5.17& 13.20 \\
                      & K+Y & 39.63 & 74.71& 56.44 \\
                      & O+K+Y & 42.45 & 79.88& 69.64 \\
UMCP & O+K+Y & \textbf{3.62}& \textbf{5.38}& \textbf{27.10} \\
\bottomrule [1pt]
\end{tabular}
}
\end{table}

\subsection{Ablation Study}
To evaluate the contribution of each component to orientation estimation, ablation experiments are conducted, as summarized in Tab.~\ref{tab:ab1}. YOLOv12-based model serves as the baseline, achieving an AAE of $9.05^\circ$ for orientation estimation. This baseline employs standard one-hot supervision for discrete orientation estimation, which does not explicitly model angular continuity. By incorporating the OFEM, the AAE is reduced to $8.63^\circ$, indicating that orientation feature enhancement leads to more informative representations for orientation estimation. Furthermore, introducing the KL divergence loss further reduces the AAE to \textbf{$7.76^\circ$}. This improvement can be attributed to the use of KL divergence loss with circularly smoothed labels, which explicitly models angular periodicity and alleviates the discontinuity inherent in one-hot supervision. Overall, the results demonstrate that both OFEM and the proposed KL loss make complementary contributions to accurate orientation estimation.

\begin{table}[t]
\centering
\caption{Ablation study on orientation estimation}
\label{tab:ab1}
\resizebox{1\linewidth}{!}{%
\renewcommand{\arraystretch}{1.6}  
\setlength{\tabcolsep}{10pt} 
\begin{tabular}{c c c c c}
\toprule[1pt]
\multirow{2}{*}{Method} & \multicolumn{3}{c}{Components} & \multirow{2}{*}{AAE } \\
\cline{2-4}
 & YOLOv12-based & OFEM & KL Loss &  \\
\hline
Baseline        & $\checkmark$ &            &            & 9.05 \\
OFEM           & $\checkmark$ & $\checkmark$ &            & 8.63 \\
 UMCP      & $\checkmark$ & $\checkmark$ & $\checkmark$ & \textbf{7.76} \\
\bottomrule [1pt]
\end{tabular}
}
\end{table}

\subsection{Real Experiment}
\begin{figure}[t]
    \centering
    \includegraphics[width=\linewidth]{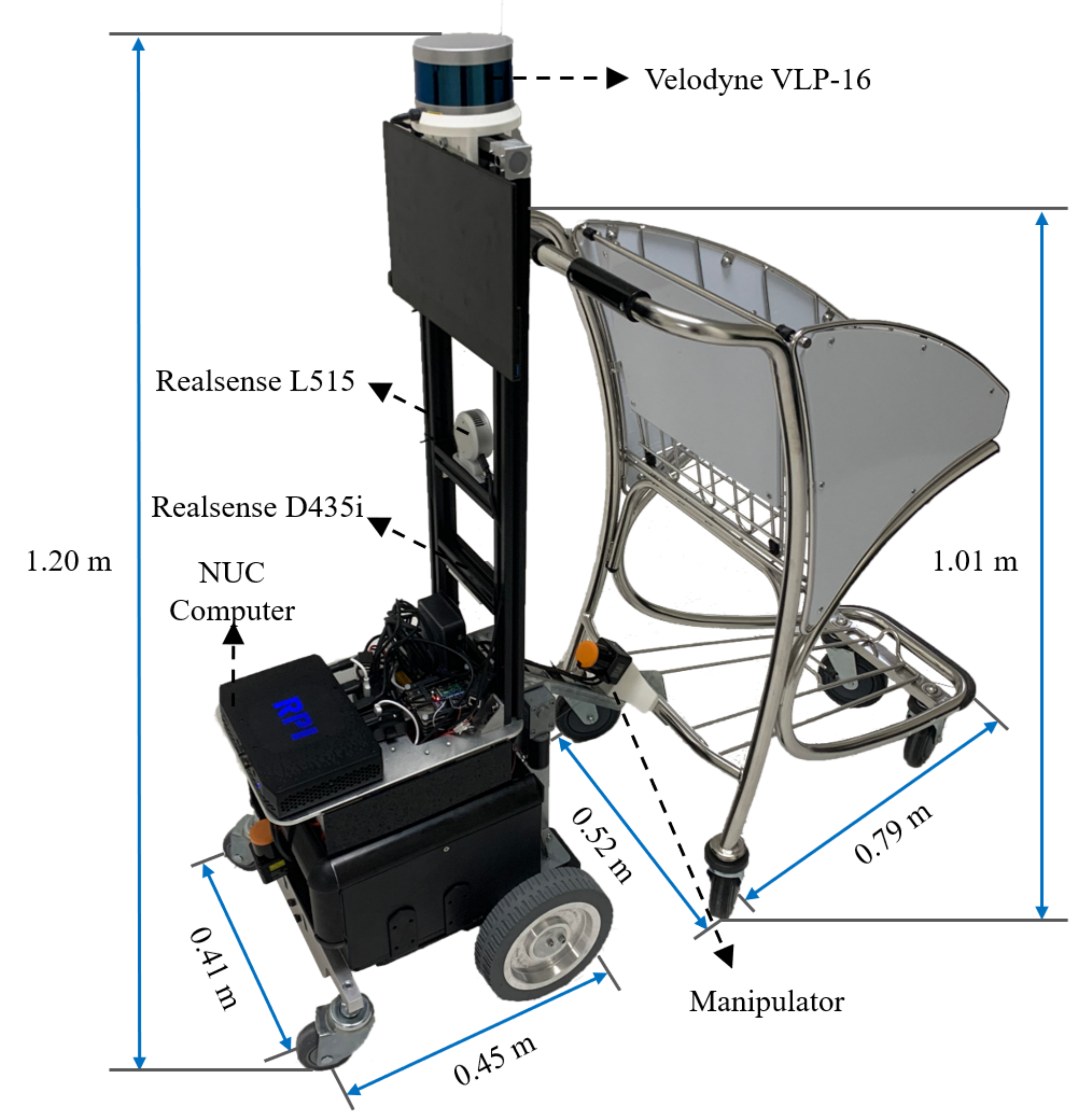}
     \caption{Experimental platform for real-world experiment.}
     \label{fig:robot}
\end{figure}
To validate the reliability of the proposed method, we conducted real-world experiments.
As shown in Fig.~\ref{fig:robot}, the proposed method is implemented on an autonomous robot designed for airport luggage trolley collection. The robot measures 0.45 m × 0.41 m × 1.20 m, while the luggage trolley has dimensions of 0.79 m × 0.52 m × 1.01 m. The robot is equipped with a Velodyne VLP-16 LiDAR, an Intel Realsense L515, an Intel RealSense D435i RGB-D camera and an onboard computer for perception, planning and control. A customized manipulator is integrated into the platform to enable efficient luggage trolley collection. The proposed algorithms are implemented within the Robot Operating System (ROS) framework and run in real time on the onboard computer, which is powered by an Intel Core i7-1165G7 CPU and an NVIDIA GeForce RTX 2060 GPU.

\begin{figure*}[htbp]
    \centering
    \includegraphics[width=0.98\linewidth]{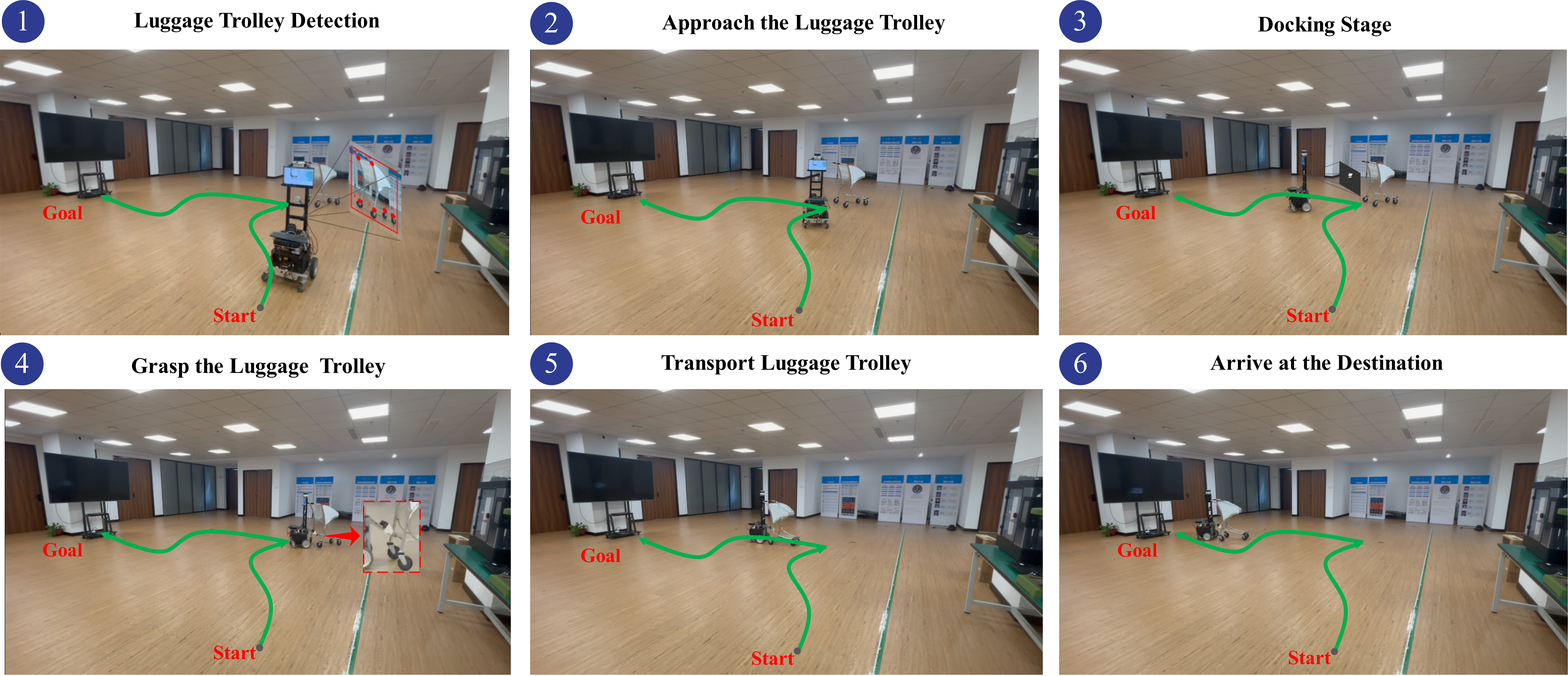}
   \caption{Sequential snapshots of our system performing a real-world luggage trolley collection task. The robot observes the pose of the  luggage trolley, approaches and docks with it, grasps the luggage trolley, transports it and finally arrives at the designated target location. Gray dot denotes the robot's initial position. The green curve represents the robot's trajectory. }
     \label{fig:real_collection_task}
\end{figure*}

As shown in Fig.~\ref{fig:real_collection_task}, a sequence of snapshots illustrates the luggage trolley collection task performed by the airport luggage trolley robot. In the initial stage, the robot estimates the pose of the luggage trolley and plans a path from its starting position to a position behind the trolley before approaching it. In the second stage, the robot performs precise perception to detect the planar surface at the rear of the luggage trolley and refines its pose estimation accordingly. Subsequently, the robot executes the grasping operation and transports the luggage trolley to the designated target location. A demonstration video of the proposed method can be found online.\footnote{\href{https://www.youtube.com/watch?v=lzvwcGc9ScY}{https://www.youtube.com/watch?v=lzvwcGc9ScY}}

 We also compared it against two state-of-the-art approaches: Xiao's method and HPPS. All comparative experiments employed the publicly available trained models provided in their respective papers. The Mean Absolute Error (MAE) and Root Mean Square Error (RMSE) were adopted as evaluation metrics, defined as follows:

\begin{align}
\text{MAE} &= \frac{1}{n}\sum_{i=1}^{n}\left(|x_i - \hat{x}_i| + |y_i - \hat{y}_i|\right), \label{eq:mae} \\
\text{RMSE} &= \sqrt{\frac{1}{n}\sum_{i=1}^{n}\left[(x_i - \hat{x}_i)^2 + (y_i - \hat{y}_i)^2\right]}. \label{eq:rmse}
\end{align}

\begin{figure}[t]
    \centering
        \includegraphics[width=\linewidth]{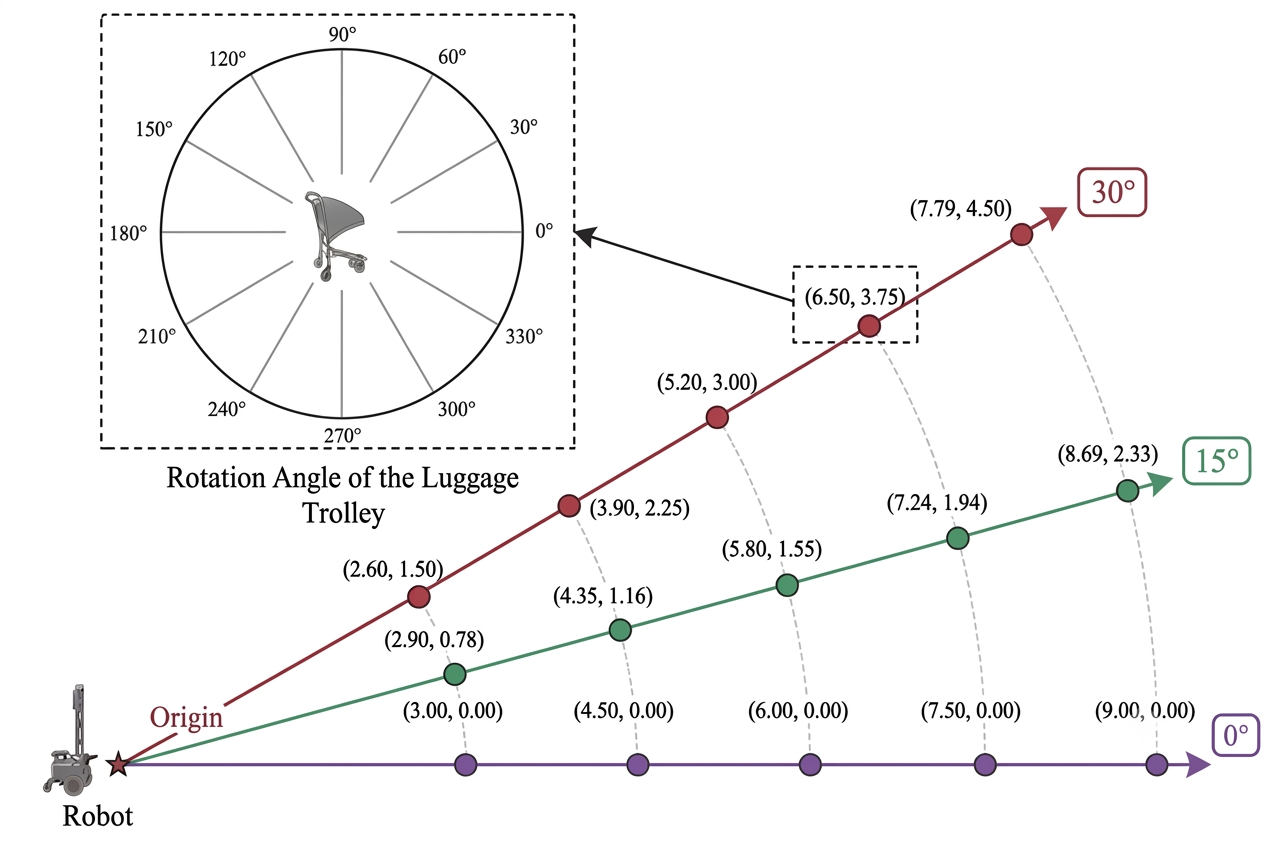}
     \caption{Different initial states of the luggage trolleys. The  point is the pose of the robot. The
    dot represents the different initial poses of the luggage trolley. Each dot is similar to a polar coordinate,
    with different distances and rotation angles, as shown in the semicircle shape. The luggage trolley has
    twelve rotation angles at each  point, as shown in the square.}
     \label{fig:real_exe}
\end{figure}

\begin{figure}[t]
    \centering
    \includegraphics[width=\linewidth]{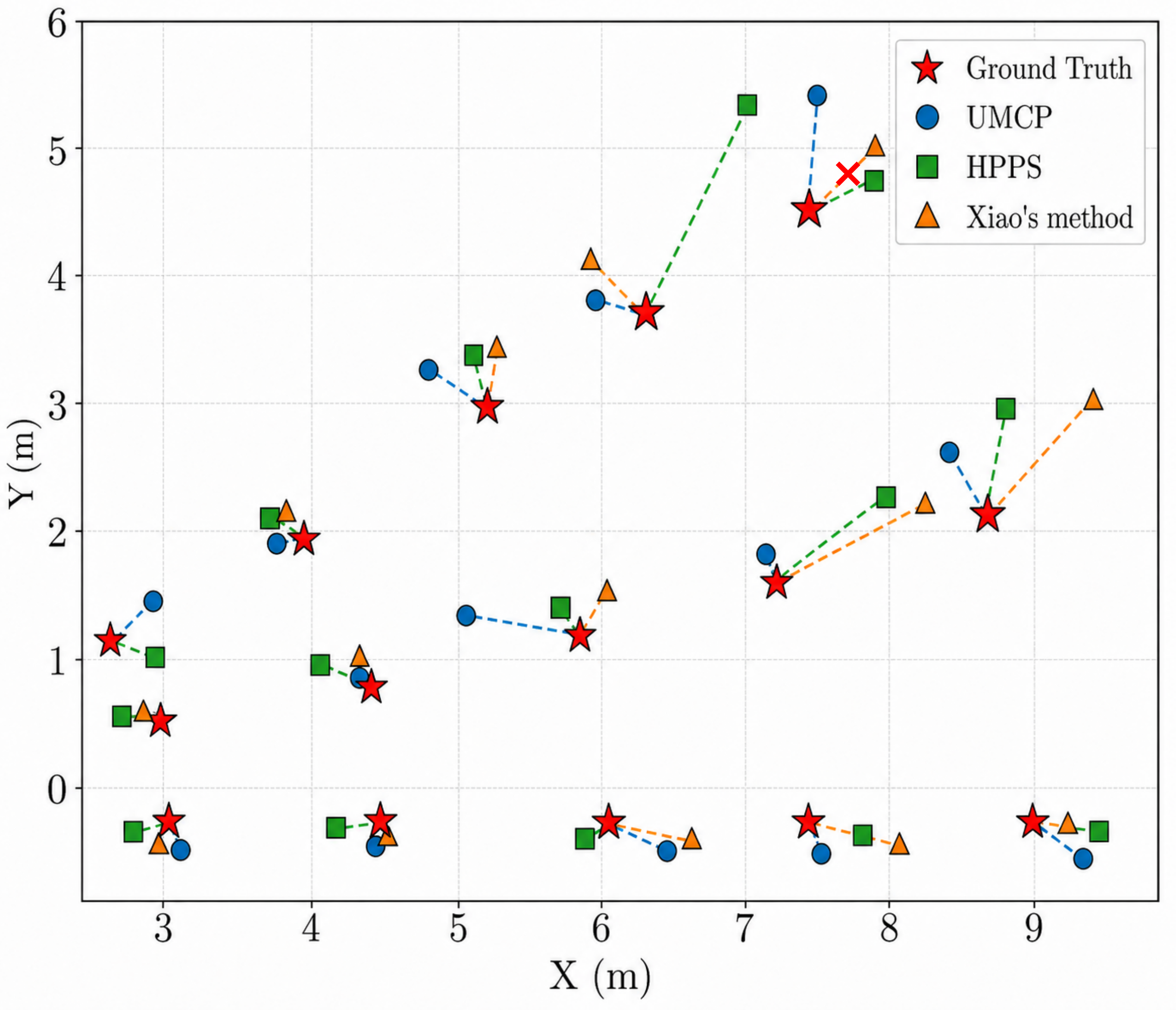}
     \caption{This figure presents the mean center point estimates obtained by each method over all angular orientations at individual test positions.}
     \label{fig:real_data}
\end{figure}

The experimental platform utilized an Intel RealSense D435i depth camera mounted at a height of 0.57 m above the ground. All three methods were deployed and tested on an NVIDIA RTX 3070 GPU. The experimental scenario was set up in an indoor environment, as illustrated in Fig.~\ref{fig:real_exe}. The robot position was defined as the coordinate origin, with the camera facing direction as the positive $x$-axis and the robot's left side as the positive $y$-axis. Given the symmetry of the scene, measurements were conducted within a single quadrant only.
The detailed experimental configuration is as follows: starting from the $x$-axis direction as $0^\circ$, measurement directions were established at $15^\circ$ intervals in the counter-clockwise direction; five test points were selected along each direction with an inter-point spacing of 1.5 m and the initial measurement position was 3 m from the robot. At each test point, the luggage trolley was rotated in place, with data sampled every $30^\circ$. The luggage trolley moved out of the robot's field of view when the rotation angle exceeded $45^\circ$. 
Tab.~\ref{tab:mae_rmse} presents the quantitative results of the three methods across all test positions. The experiments demonstrate that UMCP achieves the best overall performance, followed by HPPS. Furthermore, Fig.~\ref{fig:real_data} illustrates the mean center-point estimates at each test location, with connecting lines to the ground truth for comparative analysis, where red crosses indicate locations at which Xiao's method fails to produce valid measurements. To ensure result reliability, outlier values were excluded during the experiments. The results reveal that the localization errors of all three methods remain relatively small when the target distance is less than 4 m; however, the error fluctuation increases significantly beyond this range. Moreover, the error is minimal near the $0^\circ$ direction and the localization errors of all methods increase as the viewing angle deviates from this direction.

\begin{table}[htbp]
\centering
\caption{Comparison of Localization Errors Between Different Methods}
\label{tab:mae_rmse}
\renewcommand{\arraystretch}{1.5}
\setlength{\tabcolsep}{15pt}
\begin{tabular}{c c c}
\toprule[1pt]
Method & MAE & RMSE \\
\hline
Xiao's method~\cite{xiao2022robotic} & 0.6361 & 0.8546 \\
HPPS~\cite{sun2025hp} & 0.4084 & 0.5478 \\
UMCP & \textbf{0.3365} & \textbf{0.5442} \\
\bottomrule[1pt]
\end{tabular}
\end{table}

\section{Conclusions and Future Work}
In summary, this article presents a lightweight UMCP framework that achieves competitive performance with low deployment cost and reduced computational burden, making it well-suited to resource-constrained robotic platforms. By unifying luggage trolley detection, keypoint detection and orientation estimation within a single collaborative framework, the proposed method enables more efficient feature sharing and reduces redundancy in cascaded multi-stage perception pipelines. Experimental results further demonstrate that the proposed UMCP framework provides effective and reliable overall perception performance, particularly in cluttered and partially occluded scenarios. In future work, we will extend the network to a broader range of objects and application scenarios.

\bibliographystyle{IEEEtran} 
\bibliography{refs} 
\end{document}